\documentclass[11pt]{article}       
\usepackage{geometry}               
\usepackage{graphicx}
\usepackage{caption}
\usepackage{subcaption}
\usepackage{comment}

\usepackage{authblk}
\usepackage{amsmath} 
\usepackage{amssymb}  
\usepackage{amsthm}  
\usepackage{bm}
\usepackage{lipsum}
\usepackage{color}
\usepackage{multirow}
\usepackage{array}
\usepackage{cite}
\usepackage{hyperref}

\usepackage{makecell}

\title{Deep Reinforcement Learning for the Design of Metamaterial Mechanisms with Functional Compliance Control}
\usepackage{authblk}

\author[1]{Yejun Choi\thanks{co-first authors.}}
\author[2]{Yeoneung Kim\thanks{co-first authors.}}
\author[1]{Keun Park\thanks{corresponding author: kpark@seoultech.ac.kr}}

\affil[1]{Department of Mechanical System Design Engineering, Seoul National University of Science and Technology, Seoul 01811, Republic of Korea}
\affil[2]{Department of Applied Artificial Intelligence, Seoul National University of Science and Technology, Seoul 01811, Republic of Korea}

\date{}
\providecommand{\keywords}[1]{\textbf{Key words.} #1}

\begin{document}
\maketitle

\pagestyle{myheadings}
\thispagestyle{plain}

\begin{abstract}
Metamaterial mechanisms are micro-architectured compliant structures that operate through the elastic deformation of specially designed flexible members. This study develops an efficient design methodology for compliant mechanisms using deep reinforcement learning (RL). For this purpose, design domains are digitized into finite cells with various hinge connections, and finite element analyses (FEAs) are conducted to evaluate the deformation behaviors of the compliance mechanism with different cell combinations. The FEA data are learned through the RL method to obtain optimal compliant mechanisms for desired functional requirements. The RL algorithm is applied to the design of a compliant door-latch mechanism, exploring the effect of human guidance and tiling direction. The optimal result is achieved with minimal human guidance and inward tiling, resulting in a threefold increase in the predefined reward compared to human-designed mechanisms. The proposed approach is extended to the design of a soft gripper mechanism, where the effect of hinge connections is additionally considered. The optimal design under hinge penalization reveals remarkably enhanced compliance, and its performance is validated by experimental tests using an additively manufactured gripper. These findings demonstrate that RL-optimized designs outperform those developed with human insight, providing an efficient design methodology for cell-based compliant mechanisms in practical applications.
\end{abstract}


\section{Introduction}

Metamaterials are generally defined as artificially designed architectures exhibiting extraordinary physical properties beyond those found in natural or chemically synthesized materials~\cite{liu2011metamaterials}. While initial research primarily focused on alterations in optical or acoustic properties~\cite{shalaev2005negative,valentine2008three,ma2016acoustic}, the scope of metamaterials has since expanded to include mechanical properties, leading to the emergence of mechanical metamaterials~\cite{zadpoor2016mechanical,qi2022recent,barchiesi2019mechanical}. Recent advances in additive manufacturing (AM) technology have enabled the development of various mechanical metamaterials with complex microscale architectures. These advancements have opened up numerous possibilities across diverse fields, facilitating the creation of materials with negative material properties~\cite{lakes2008negative,ai2017metamaterials,saxena2016three}, extreme mechanical properties~\cite{kadic2012practicability,zheng2014ultralight,davami2015ultralight}, and exceptional thermal properties~\cite{vemuri2014guiding,you2021design,sha2021robustly}.

Micro-architectured mechanical metamaterials have also been utilized in designing compliant mechanisms, known as \textit{metamaterial mechanisms}~\cite{ion2016metamaterial}. A compliant mechanism is a mechanical device that operates through the elastic deformation of its flexible members~\cite{ling2020kinetostatic}. Unlike traditional mechanisms, which require the assembly of multiple components, compliant mechanisms eliminate the need for subsequent assembly and avoid joint clearance issues~\cite{cuellar2018additive}. Among various mechanical metamaterials, re-entrant auxetic structures with tunable Poisson’s ratio have been employed to develop compliant mechanisms~\cite{ren2018auxetic}, including soft robot actuators~\cite{mark2016auxetic} and jointless hinges~\cite{hedayati20213d}. Additionally, three-dimensional chiral metamaterials have been explored to achieve compliant mechanisms based on the compression-torsion deformation characteristics~\cite{frenzel2017three,wu2019mechanical,ji2024design}.

While compliant mechanisms have traditionally been designed using human insights or finite element analysis (FEA), topology optimization (TO) has also been employed to achieve structurally optimized geometries~\cite{zhu2020design}. The TO method aims to minimize or maximize structural compliance under given load conditions~\cite{sigmund2013topology} and has been effectively utilized to enhance structural efficiency in conjunction with AM processes~\cite{zegard2016bridging,mass2017topology,kim2021multiscale}. This method has been particularly applied to design compliant mechanisms with desired shape morphing~\cite{kumar2020topology} or enlarged deformation areas~\cite{kumar2021topology}. Further studies have explored integrating the TO method with machine learning (ML) technology, paving the way for more advanced and efficient design processes~\cite{shin2023topology}. However, this integrated approach requires high computational loads due to the significant iterative calculations involved in both TO and ML technologies.

In this study, we propose an efficient and straightforward application of ML to design a compliant mechanism without additional computation for TO. For this purpose, we digitize the design domain into a finite number of cells with different hinge connections. The framework of reinforcement learning (RL) is then employed to obtain optimal compliant mechanisms that satisfy desired functional requirements by maximizing the reward in the sequential selection of cell designs. RL is recognized as an effective ML method for identifying policies that maximize rewards in complex engineering problems~\cite{arulkumaran2017deep} and has been applied to develop mechanical metamaterials~\cite{song2024artificial}. Applications of RL includes tuning of material properties~\cite{xu2021machine,sui2021deep,brown2023deep}, shape optimization~\cite{viquerat2021direct}, topology optimization~\cite{brown2022deep}, and thermoresponsive 4D printing~\cite{mohammadi2024sustainable}. These studies have demonstrated considerable efficacy in addressing complex optimization problems inherent in the characterization and synthesis of materials using the RL method.

While previous studies applied RL to the optimal design of static structures, this study extends its application to the design of compliant mechanisms. A door-latch mechanism is considered as a target compliant mechanism to achieve the linear motion of a latch through rotational deformation applied to the axle. The design domain is digitized to incorporate various types of unit cells, whose deformation behaviors are investigated by structural FEAs. The FEA results are then effectively exploited through a carefully crafted reward function, and a deep Q-learning framework, one of the well-established RL algorithms, is applied to determine the optimal compliant mechanism. In the learning procedure, the effects of human guidance and cell-tiling direction are investigated in terms of optimization performance. Furthermore, the RL-based design approach is extended to design of a soft gripper mechanism, with additional discussion on the effect of hinge connections. The optimally designed compliant mechanisms are then additively manufactured and experimentally validated, demonstrating that the proposed approach provides a useful design methodology for compliant mechanisms with an optimized combination of digitized cells. This methodology not only enhances the efficiency of the design process but also ensures the creation of compliant mechanisms with precise and reliable deformation characteristics suitable for practical applications.





\section{Methods}\label{sec:method}
\subsection{Design of Compliant Mechanism}
\begin{figure}[h]
  \centering
  \includegraphics[width=.8\textwidth]{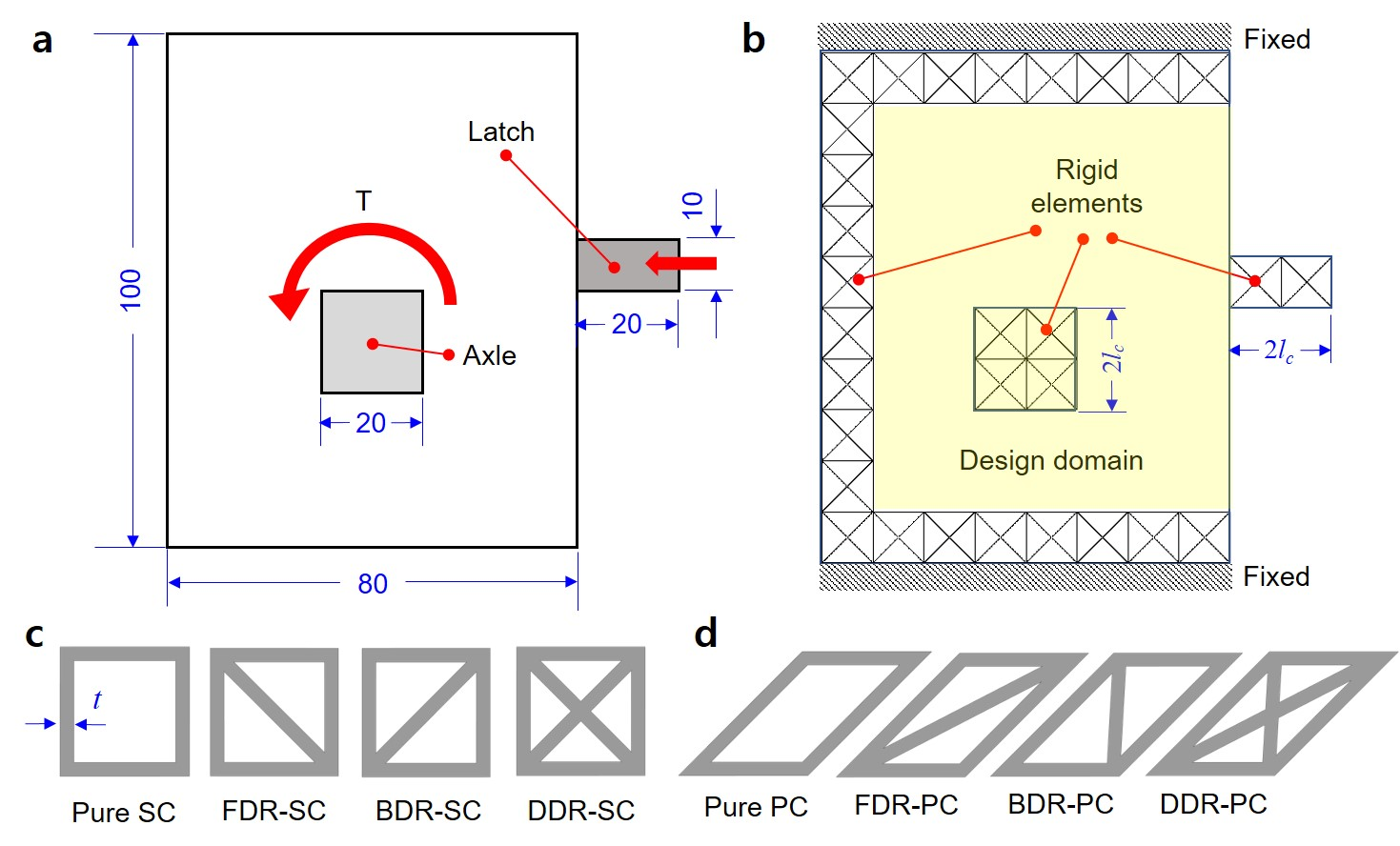}
  \caption{Design of a compliant door-latch mechanism: (a) basic design configuration (unit: mm), (b) selection of rigid elements, (c) square cells with different diagonal reinforcements, and (d) parallelogram cells with different diagonal reinforcements.}
  \label{fig:compliant}
\end{figure}

A door-latch mechanism serves as an example of a compliant mechanism using various types of cells, which was developed to replace the traditional door-latch mechanism consisting of several mechanical components~\cite{ion2016metamaterial}. Figure~\ref{fig:compliant}a depicts the configuration of a compliant door-latch mechanism within a rectangular domain measuring $80 \times 100$ mm. The mechanism comprises a rectangular latch, measuring $20 \times 10$ mm, attached on the right side of the domain. A square axle with dimensions of $20\times20$ mm is situated inside the rectangular domain. This mechanism is designed to induce horizontal movement of the latch through rotational deformation applied to the axle.

Figure~\ref{fig:compliant}b presents the initial design configuration of the compliant door-latch mechanism, where several rigid cells are predefined. A series of square cells with double-diagonal reinforcement is arranged for the axle, latch, and three boundaries (i.e., top, bottom, and left edges). The size of each cell ($l_c$) and wall thickness ($t$) are set to $10$ and $1.2$ mm, respectively. These square cells with double-diagonal reinforcement are intended to act as rigid elements due to their high stiffness compared to other cells with less diagonal reinforcement. The remaining region corresponds to the design domain where various cell types are assigned to induce the desired motion.

To fill the design domains in a digitized manner, two types of cells, a square cell (SC) and a parallelogram cell (PC) are defined. Figure~\ref{fig:compliant}c demonstrates four subtypes of SCs with different diagonal reinforcement: (i) pure SC without reinforcement, (ii) SC with forward-diagonal reinforcement (FDR-SC), (ii) SC with backward-diagonal reinforcement (BDR-SC), and (iv) SC with double-diagonal reinforcement (DDR-SC). This varying reinforcements aim to diversify the deformation behaviors, thus realizing the desired motion. Similarly, four subtypes of PCs with different diagonal reinforcement can be defined: (i) pure PC without reinforcement, (ii) PC with forward-diagonal reinforcement (FDR-PC), (ii) PC with backward-diagonal reinforcement (BDR-PC), and (iv) PC with double-diagonal reinforcement (DDR-PC). The relevant deformation behaviors of these cells are numerically investigated through FEA, and the 
relevant results are discussed in Section~\ref{subsec:deformation}.

\subsection{Materials}
For AM of compliant door-latch structures, we utilizes thermoplastic polyurethane (TPU) filaments with a diameter of 1.75 mm (eTPU-95A, Shenzhen ESUN Industrial Co. Ltd., China), considering its high elongation and flexibility. The TPU filament has a density of 1.21 g/cm$^3$, an elastic modulus of 24.1 MPa, and a Poisson’s ratio of 0.39. Additively manufactured parts using this filament are known to have shore hardness of 95A, tensile strength of 35 MPa, and tensile elongation of 800\%\cite{tpu}. 

Additional components for experimental equipment were additively manufactured using polylactic acid (PLA) filaments (PLA-i21, Cubicon Inc., Korea). This material has a density of 1.24 g/cm$^3$ and an elastic modulus of $3.5$ GPa. Additively manufactured parts using this filament exhibit a tensile strength of $53$ MPa and tensile elongation of $7.2\%$\cite{pla}. Considering that its elastic modulus is $145$ times higher than that of TPU (24.1 MPa), the additively manufactured parts using this filament can be regarded as rigid bodies when they are assembled with a TPU-based compliant structure.

\subsection{Finite Element Analysis}\label{subsec:fea}
Finite element analyses (FEAs) are conducted to investigate the structural deformation behaviors of various unit cells as shown in Figures \ref{fig:compliant}c and \ref{fig:compliant}d. The FEAs are performed using ANSYS $\text{Workbench}^\text{®}$ software (ANSYS Inc., USA). A two-dimensional (2D) plain strain model is adopted for the FEA to efficiently represent the geometric characteristics of the unit cells. The 2D analysis domains are discretized using an eight-node quadrilateral element with a $0.3$ mm mesh size.

\begin{figure}[h]
  \centering
  \includegraphics[width=.8\textwidth]{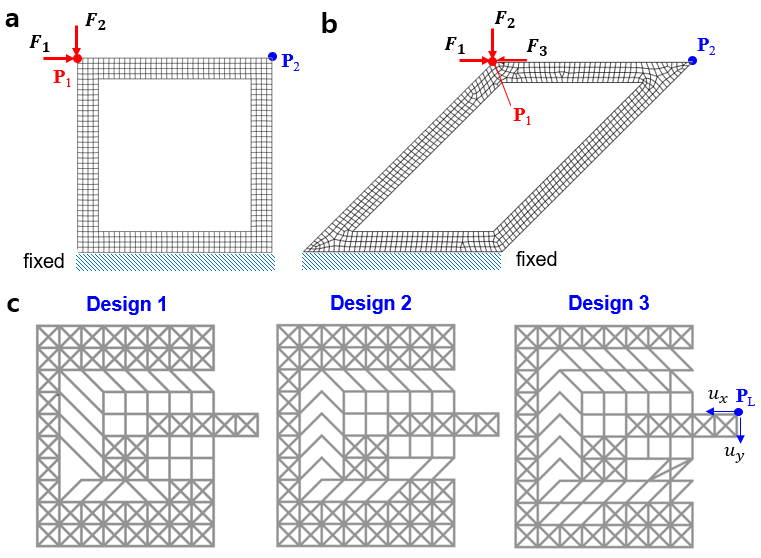}
  \caption{Various FEA models: (a) square unit cell, (b) parallelogram unit cell, (c) three design cases for the door-latch structure (Designs 1-3).}
  \label{fig:structure}
\end{figure}

Figures~\ref{fig:structure}a and \ref{fig:structure}b illustrate the finite element mesh with boundary conditions for square and parallelogram unit cells, respectively. In both cases, the bottom edges of the cells are fixed, and force conditions are applied at the top-left corner points $(\textbf{P}_1)$. As illustrated in Figure~\ref{fig:structure}a, two load conditions ($\text{F}_1$ and $\text{F}_2$), horizontal and vertical forces with a magnitude of 100 N, are applied for a square cell. For a parallelogram cell, a vertical force ($\text{F}_2$) and two horizontal forces with the same magnitude and opposite direction ($\text{F}_1$ and $\text{F}_3$) are applied considering its asymmetric geometry, as depicted in Figure~~\ref{fig:structure}b. While these figures represent the pure cells without reinforcement, all reinforced cells were discretized similarly and have identical boundary conditions.

Figure~\ref{fig:structure}c illustrates three design cases for the door-latch structure, where the design domain is filled with various arrangements of square and parallelogram cells. For computational efficiency, this structure is discretized using one-dimensional (1D) beam elements with a mesh size of $0.1$ mm. FEAs are conducted for these structures with the boundary conditions described in Figures~\ref{fig:compliant}a and~\ref{fig:compliant}b, wherein the top and bottom edges are fixed, and a counterclockwise torque ($T$) is applied at the axle. Consequently, the latch is expected to move in the left direction, and the corresponding displacements of the latch tip ($\textbf{P}_\text{L}$) are measured to assess the directional compliance of the designed structures.

\subsection{Experiments}\label{subsec:exp}
\subsubsection{Experimental Setup}
For experimental validation of the compliant door-latch structures, a fixture part with a size of $100 \times 130 \times 25$ mm is designed as illustrated in Figure 3a. The additively manufactured door-latch structure is then assembled into this fixture, and a door handle is assembled onto the axle of the door-latch. A downward vertical force is applied to the tip of the handle, resulting in rotational torque on the axle. A rectangular hole with a width of 15 mm is designed near the latch, allowing for 2 mm gap distances along the upper and lower directions to accommodate the vertical deflection of the latch.

\subsubsection{Additive manufacturing}
The designed compliant structures for experimental validations are additively manufactured using a fast-filament fabrication (FFF) type 3D printer (Cubicon Neo-A22C, Cubicon Inc., Korea). This printer is equipped with a temperature-controlled printing chamber, maintaining a temperature of 40$^\circ \text{C}$ throughout the printing process. The diameter of the extrusion nozzle is 0.4 mm, and the layer thickness is configured to 0.2 mm.

Various door-latch structures with different cell designs are additively manufactured using TPU filaments. For the AM of TPU material, the nozzle and bed temperatures are set to 230$^\circ \text{C}$ and 65$^\circ \text{C}$, respectively. The fixture and door handle components, which require higher rigidity than the door-latch structure, are fabricated using PLA filaments. For the AM of PLA material, the nozzle and bed temperatures are set to 210$^\circ \text{C}$ and 60$^\circ \text{C}$, respectively. The printing speed is set to 60 mm/s for both filaments. Figure 3b demonstrates the experimental setup in which the additively manufactured components are assembled.

\begin{figure}[h]
  \centering
  \includegraphics[width=.8\textwidth]{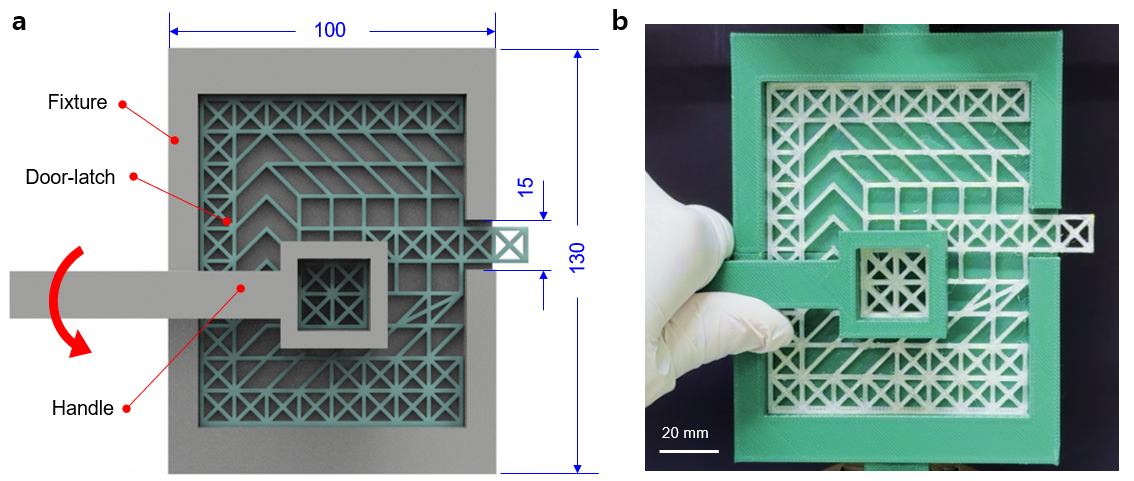}
  \caption{Experimental configuration for compliant door-latch structures: (a) schematics of the experimental setup, (b) assembled components for additively manufactured door-latch mechanism.}
  \label{fig:exp_con}
\end{figure}

\subsection{Compliant Mechanism Design via deep RL}\label{sec:comp_deeprl}
For optimal design of compliant mechanisms, the deep RL algorithm involves transforming the design problem into a Markov decision process (MDP) and employing a dueling Deep Q-Network (DQN) for enhanced learning. Detailed methodologies are explained in the following subsections.

\subsubsection{Q-learning for Markov decision process}
A discrete Markov decision process consists of state space ($\mathcal{S}$), an action space ($\mathcal{A}$), a transition probability function, a reward function ($r:\mathcal{S} \times \mathcal{A} \rightarrow \mathbb{R}$), and a discount factor $\gamma \in [0,1]$. Since we consider a discrete MDP, $\mathcal{S}$ and $\mathcal{A}$ are assumed to be finite. An agent at state $s_t \in \mathcal{S}$ in the MDP sequentially chooses an action $a_t \in \mathcal{A}$, receives a reward given by $r(s_t,a_t)$, and moves to the state $s_{t+1}$ according to the transition probability $\mathrm{Pr}(s_{t+1}|s_t,a_t)$, which describes the probability distribution of the next state $s_{t+1}$ given the current state and action pair $(s_t,a_t)$.

The goal of an agent in an MDP is to find an optimal policy function $\pi: \mathcal{S} \rightarrow \mathcal{A}$ that maximizes the cumulative reward defined as
\begin{equation}
\mathbb{E}_{\pi} \left[\sum_{t=1}^H \gamma^t r(s_t, a_t) \right],
\end{equation}
where $H$ represents the number of states that an agent has observed, often called the length of an episode.

The discount factor $\gamma$ plays a crucial role in determining the present value of future rewards; a discount factor close to 1 indicates that future rewards are nearly as valuable as immediate rewards, whereas a smaller discount factor places more emphasis on immediate rewards. This factor contributes to ensuring the convergence of value functions by Banach's fixed point theorem and balancing short-term and long-term returns in RL~\cite{sutton2018reinforcement}.

Given a policy denoted by $\pi$, the value functions are defined as follows:
\begin{equation}
V^\pi(s) := \mathbb{E}_{\pi} \left[\sum_{t=1}^H \gamma^t r(s_t, a_t) \mid s_0=s\right],
\end{equation}
and
\begin{equation}
Q^\pi(s,a):= \mathbb{E}_{\pi} \left[\sum_{t=1}^H \gamma^t r(s_t, a_t) \mid s_0=x,a_0=a \right],
\end{equation}
where trajectories are generated following the policy $\pi$. Then the optimal policy $\pi^*$ satisfies
\begin{equation}
V(s):=V^{\pi^*}(s)=\max_{\pi} V^\pi (s) \quad \text{and}\quad Q(s,a):=Q^{\pi^*}(s,a)=\max_\pi Q(s,a)
\end{equation}
for any $s$, $a$ and we deduce Bellman's optimality equation:
\begin{equation}
V(s) = \max_{a \in \mathcal{A}} \left[ r(s, a) + \gamma \sum_{s' \in \mathcal{S}} \mathrm{Pr}(s'|s, a) V(s') \right],
\end{equation}
and
\begin{equation}
Q(s, a) = r(s, a) + \gamma \sum_{s' \in \mathcal{S}} \mathrm{Pr}(s'|s, a) \max_{a' \in \mathcal{A}} Q(s', a').
\end{equation}
These equations provide the foundation for RL algorithms such as value iteration and policy iteration. In particular, we focus on Q-learning~\cite{watkins1992q}, an off-policy RL algorithm, for our task. In Q-learning, the Q-function is updated until it converges for all state-action pairs following:
\begin{equation}
Q(s_t,a_t) \leftarrow Q(s_t,a_t)+\alpha (r(s_t,a_t)+\gamma \max_{a'\in\mathcal{A}} Q(s_{t+1},a')-Q(s_t,a_t)),
\end{equation}
where $\alpha>0$ is given. 

\subsubsection{Transformation of design problem into MDP}
As illustrated in Figure~\ref{fig:cell_place}c, twelve possible unit cells will be placed in the design domain to obtain the desirable mechanism. Instead of considering all possible configurations ($= 12^\text{H}$), this problem is regarded as a sequential decision-making problem or MDP with an RL algorithm to let the placement policy interact with the environment. Through this interaction, an agent seeks the best configuration with the desirable mechanism. 

\begin{figure}[h]
  \centering
  \includegraphics[width=.8\textwidth]{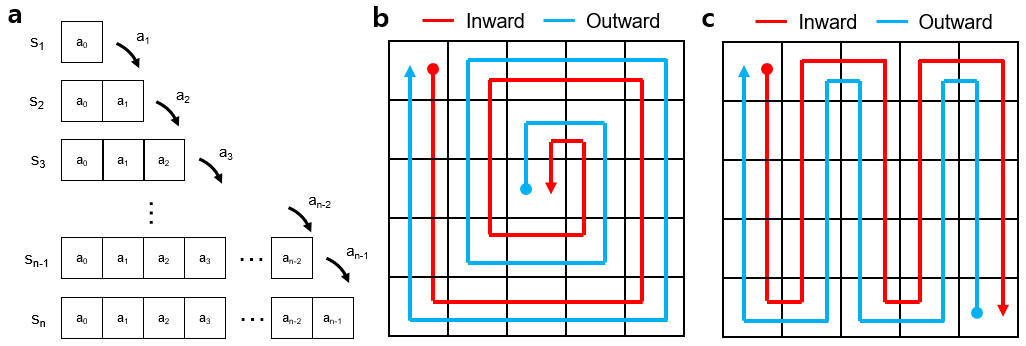}
  \caption{Conversion of mechanism design problem into sequential decision-making problem: (a) conversion to a sequential decision-making problem by unfolding, (b) spiral tiling, (c) zigzag tiling.}
  \label{fig:placement_methods}
\end{figure}

With the fixed horizon $H$ that represents the number of cells, let the state $s_t$ denote the configuration after placing $t$ cells. The agent chooses an action $a_t \in \mathcal{A}$ consisting of twelve unit cells to reach the next state $s_{t+1}$. For the reward function, we set $r_i=0$ for $i\in \{1,...,H-1\}$ and $r_H$ is given as the deformation computed via FEA. To transform the design of the compliant mechanism problem into an MDP problem, we unfold the cells from inside to outside or in a reverse direction as shown in Figure~\ref{fig:placement_methods}a. At $t=0$, an agent chooses an arbitrary action $a_0$ to reach $s_1$.

It is also worth noting that there are various ways of placing cells sequentially. Depending on the desired mechanism, the direction of sequential decision-making (i.e., the tiling direction) must be carefully determined. Figure~\ref{fig:placement_methods}b illustrates a spiral tiling method, which offers both the inward and outward directions. Figure~\ref{fig:placement_methods}c shows a zigzag tiling method, also with inward and outward directions. The selection of the tiling direction is relevant to the deformation mechanism of a compliant structure. For example, in the case of the door latch mechanism, the spiral tiling is more appropriate since the deformation mechanism is initiated by the rotation of the axle, as illustrated in Figure~\ref{fig:compliant}a.


\subsubsection{Dueling DQN}
\begin{figure}[h]
  \centering
  \includegraphics[width=0.8\textwidth]{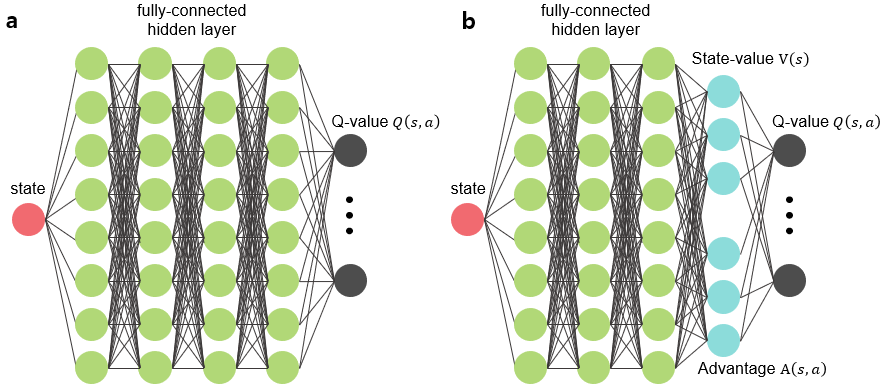}
  \caption{The neural network architectures: (a) DQN, (b) Dueling DQN.}
  \label{fig:archi}
\end{figure}

In this study, the dueling DQN algorithm~\cite{wang2016dueling} is employed to enhance the traditional DQN method, where the Q-function is approximated by a deep neural network. Whereas the traditional DQN uses a single neural network to estimate Q-values directly, dueling DQN divides the neural network into two separate streams: one for estimating the state value function $V$ and another for estimating the advantages for each action in $\mathcal{A}$. This architecture allows the model to independently learn the value of being in a particular state and the advantage of taking each action, which is then combined to compute the Q-values as illustrated in Figure~\ref{fig:archi}b. In contrast, the standard DQN architecture takes the state vector as input and outputs Q-values for each action $a$, as shown in Figure~\ref{fig:archi}a.

This enhanced feature of the dueling DQN algorithm allows us to handle situations where the value of actions varies significantly across states in the cell placement problem. The potential advantages of using dueling DQN are summarized as follows: (i) reduced variance - by decoupling the estimation of state values and action advantages, dueling DQN can reduce the variance of Q-value estimates. This reduction can lead to more stable training and faster convergence, (ii) improved generalization - by explicitly modeling the value and advantage components separately, Dueling DQN has the potential to generalize better across different states and actions. This is particularly beneficial in environments with sparse rewards or complex dynamics,  (iii) performance: empirical studies have demonstrated that Dueling DQN often surpasses traditional DQN in terms of sample efficiency and final performance. This advantage is especially notable in environments with large state spaces or when the advantage of actions varies significantly across states.

The neural network model used for our experiment consists of five fully connected (FC) layers, each with a varying number of neurons. 
The dueling DQN architecture processes the agent's state input through a sequence of FC layers with 128, 256, 512, 256, and 128 neurons, respectively. Following these layers, the network splits into two streams: one for the advantage function and one for the value function. The advantage stream consists of two FC layers with 64 and 12 neurons, while the value stream includes two FC layers with 64 and 1 neuron. The advantage values are adjusted by subtracting their mean, and combined with the state value to produce the final Q-values for each action. The action associated with the highest Q-value is selected, determining the next state and subsequent observations.

\subsubsection{Learning-based design of the door-latch mechanism}
The proposed RL approach, utilizing the dueling DQN algorithm, is applied to the design of the compliant door-latch mechanism. After determining the direction for the sequential placement of unit cells, these cells are arranged to meet the desired functional requirements. Therefore, the careful design of the reward function is essential for optimizing learning efficiency. In the door-latch mechanism, the latch is expected to move horizontally as much as possible when the user rotates the axle counterclockwise as demonstrated in Figure~\ref{fig:compliant}. To meet this requirement, the reward function is formulated as follows:
\begin{equation}
r(s_t,a_t)=
\begin{cases}
    0\quad &\text{for} \quad t\in [1,H-1],\\ 
    u_x /(C+u_y^2) \quad &\text{for} \quad t=H,
\end{cases}
\end{equation}
where $u_x$ and $u_y$ denote displacement in horizontal and vertical directions. $C$ is a positive constant, which is set to $0.1$ in this problem. $H$ denotes the horizon representing the total number of cells to be placed. Therefore, maximization of this reward can be interpreted as maximizing the displacement in the horizontal direction while minimizing the displacement of the vertical direction, implying that an agent seeks a configuration where the latch is pushed inside of the domain when the handle is rotated. 

\begin{figure}[h]
  \centering
  \includegraphics[width=0.6\textwidth]{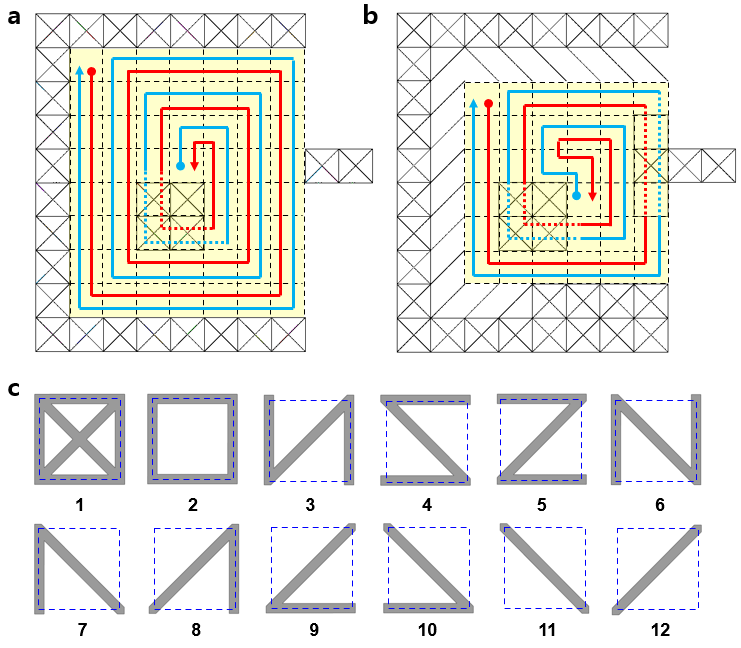}
  \caption{Cell tiling strategies for the door-latch mechanism: (a) initial configuration without a human guide, (b) initial configuration with a human guide, (c) twelve different unit cells. Here, the red and blue lines indicate the inward and outward tiling, respectively.}
  \label{fig:cell_place}
\end{figure}

To employ the RL in designing the door-latch mechanism, the design domain is digitized into 52 cells, each 10 × 10 mm in size. Two spiral tiling strategies, with the inward and outward directions, are displayed in Figure~\ref{fig:cell_place}a. In comparison to the complete void configuration, a human-guided initial configuration is also considered, as shown in Figure~\ref{fig:cell_place}b. In this configuration, cells in the outer layer are predefined by placing PC cells intended to increase rotational compliance, leaving the remaining 29 cells as the reduced design domain. This domain reduction leads to the saving of computational resources and more stable learning. Figure~\ref{fig:cell_place}c illustrates twelve different unit cells that will be selectively placed in the design domain. The effect of the human guidance and cell tiling direction will be discussed in Section~\ref{sec:result}.


To perform deep RL with dueling DQN, the discount factor, learning rate, and batch size are set to 0.99, 0.001, and 64, respectively. Adam optimizer~\cite{kingma2014adam} is used for the optimization. In the FEA to prepare the learning data, the rotational torque ($T$) is set to 5 N-m.

\section{Results}\label{sec:result}
\subsection{Deformation Behavior of Unit Cells}\label{subsec:deformation}

\subsubsection{Deflection behavior of square cells}
\begin{figure}[h]
  \centering
  \includegraphics[width=0.8\textwidth]{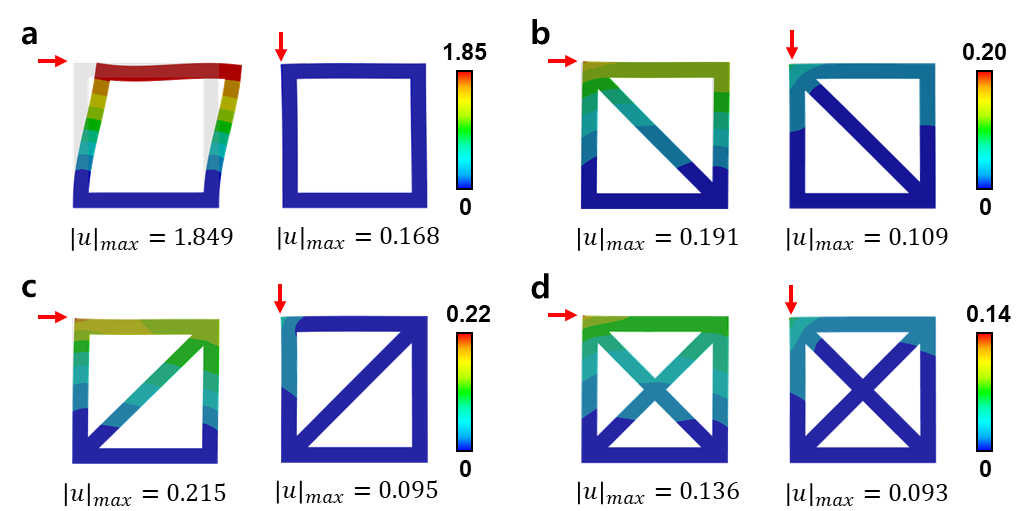}
  \caption{Deformation of various types of square unit cells (unit: mm): (a) pure SC, (b) FDR-SC, (c) BDR-SC, and (d) DDR-SC.}
  \label{fig:defor}
\end{figure}

Figure~\ref{fig:defor} represents the deformed shapes of four types of SCs under two load conditions ($\text{F}_1$ and $\text{F}_2$), wherein the color contours represent the magnitude of the displacement vector, denoted as $|u|$. Figure~\ref{fig:defor}a depicts the deformed shapes of the pure SC under two load conditions. The transverse load $(\text{F}_1)$ results in a large displacement ($|u|_{\max}$ = 1.849 mm), whereas that of the vertical load ($\text{F}_2$) is only 0.168 mm, indicating a reduction to 1/11. The other three SCs with diagonal reinforcements exhibit significantly reduced displacements under the transverse load, ranging from 0.136 to 0.215 mm. As shown in Figures~\ref{fig:defor}b and~\ref{fig:defor}c, the FDR-SC exhibits lower displacement ($|u|_{\max}$ = 0.191 mm) than the BDR-SC case ($|u|_{\max}$ = 0.215 mm) under the transverse load, while this trend is reversed in the case of the vertical load due to the direction of the diagonal reinforcement. The DDR-SC shows the smallest displacements, 0.136 and 0.093 mm for the cases of the transverse and vertical loads, respectively.

Table~\ref{table:p1} compares the directional displacement components and their corresponding magnitudes at the right-top position, denoted as $\textbf{P}_2$ in Figure~\ref{fig:structure}a. Overall, the $x$-directional displacements ($u_x$) are more significant than the $y$-directional displacements ($u_y$). Additionally, the transverse load ($\text{F}_1$) induces larger displacements than the vertical load ($\text{F}_2$). These results are attributed to the boundary condition where the bottom edge is fixed, resulting in the dominance of transverse shear deformation compared to axial deformation.

Under the transverse load, the pure SC exhibits the largest displacement, whereas the DDR-SC shows the smallest displacement. This trend is consistent with the maximum displacement observed in Figure~\ref{fig:defor}. In contrast, the vertical load case displays a slightly different trend: the horizontal displacement of DDR-SC (0.021 mm) is higher than that of BDR-SC (0.009 mm). This difference can be explained by examining the deformation contours in Figures~\ref{fig:defor}c and \ref{fig:defor}d. Specifically, the deformation of BDR-SC is concentrated on the left edge, while that of DDR-SC is uniformly distributed on the top edge. Consequently, the DDR-SC has been selected as the rigid element due to its high stiffness with uniform deformation characteristics.

\begin{table}[h]
\centering
\begin{tabular}{c|ccc|ccc }

\hline
Load    &   & \text{Transverse }$(\text{F}_1)$ &   &  & \text{Vertical} $(\text{F}_2)$ &    \\ 
\hline
Displacement (mm)    & $u_x$ & $u_y$ & $|u|$ & $u_x$ & $u_y$ & $|u|$ \\
\hline
Pure SC         & -1.753 & -0.243 & 1.770 & 0.1030 & 0.0032 & 0.1030      \\ 
FDR-SC         & -0.127 &	-0.010 &	0.127 &	-0.0377	& -0.0016	 & 0.0377	 \\
BDR-SC         & -0.128	& -0.038	& 0.134	& -0.0093	& -0.0017	& 0.0097	   \\ 
DDR-SC          &-0.066	&-0.020	&0.069	&-0.0205	&-0.0063	&0.0214	   \\ 
\hline
\end{tabular}

\caption{
Displacement results at $\textbf{P}_2$ for various SCs under different load conditions.
}
\label{table:p1}
\end{table}

\subsubsection{Deflection behavior of parallelogram cells}\label{subsec:para}
\begin{figure}[h]
  \centering
  \includegraphics[width=.5\textwidth]{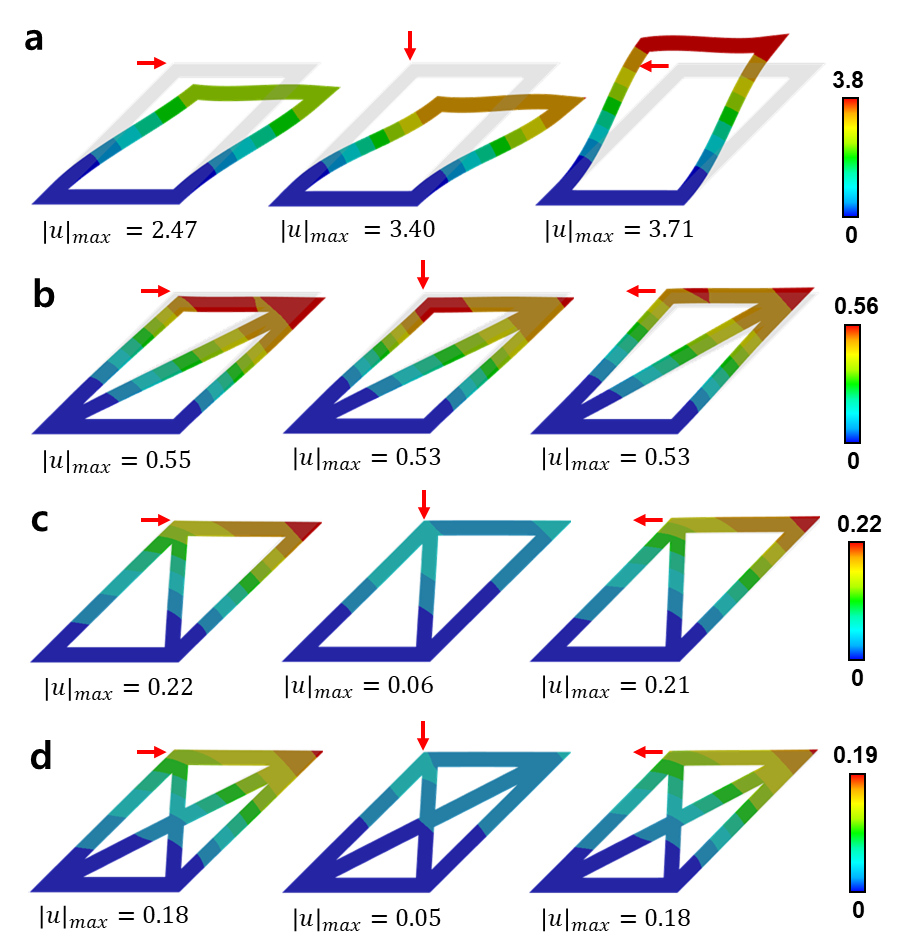}
  \caption{Deformation of various types of parallelogram unit cells (unit: mm): (a) pure PC, (b) FDR-PC, (c) BDR-PC, and (d) DDR-PC.}
  \label{fig:defor_parallel}
\end{figure}

Figure~\ref{fig:defor_parallel} represents the deformed shapes of four types of PCs under three load conditions ($\text{F}_1$, $\text{F}_2$, and $\text{F}_3$). Unlike the SC cases, two transverse loads with the same magnitude and opposite directions were applied, as described in Section~\ref{subsec:fea}. Figure~\ref{fig:defor_parallel}a depicts the deformed shapes of the pure PC, with maximum displacements for loads $\text{F}_1$, $\text{F}_2$, and $\text{F}_3$ measured at 2.47, 3.40, and 3.71 mm, respectively. These substantial displacements for all load conditions contrast with those of SC, where the vertical load induced negligible displacement (0.168 mm). This difference suggests that the PCs exhibit more flexible deformation behavior than the SCs, owing to their inclined ligaments.

Figure~\ref{fig:defor_parallel}b presents the deformed shapes of the FDR-PC under different load conditions, showing similar levels of maximum displacements ranging between 0.528 and 0.551 mm. This finding contrasts with the results of FDR-SC, where the vertical load induced lower displacements (0.109 mm) compared to the transverse load (0.191 mm), as depicted in Figure~\ref{fig:defor}b. In contrast, the BDR-PC and DDR-PC cases exhibit a different trend, with the vertical load resulting in lower displacements compared to the transverse load. These distinct deformation behaviors are attributed to the presence of vertical ligaments, which enhances the vertical stiffness of the cell.

Table~\ref{table:p2} provides a comprehensive comparison of the directional displacements and their corresponding magnitudes at the $\textbf{P}_2$ position. It is observed that the $x$-directional displacements ($u_x$) of each cell are similar in magnitude to the $y$-directional displacements ($u_y$), which is in contrast to the SC cases where the ($u_x$) values are significantly larger than the ($u_y$) values. This finding suggests that the PCs exhibit more balanced deformation behavior in both the transverse and vertical directions compared to the SCs. Additionally, when comparing the effects of load type, it is noted that the vertical load ($\text{F}_2$) induces smaller displacements than the transverse loads ($\text{F}_1$ and $\text{F}_3$) for the BDR-PC and DDR-PC cases. This indicates that the PCs demonstrate less sensitivity to load type while providing more flexible deformation behaviors compared to the SCs.

\begin{table}[h]
\centering
\begin{tabular}{c|ccc|ccc|ccc}

\hline
Load    &   & $\text{F}_1$ &   &  & $\text{F}_2$ &   & & $\text{F}_3$ & \\ 
\hline
Displacement (mm)    & $u_x$ & $u_y$ & $|u|$ & $u_x$ & $u_y$ & $|u|$  & $u_x$& $u_y$ & $|u|$\\
\hline
Pure PC         & -1.54 & -1.93 & 2.47 & -2.02 & -2.73 & 3.40       & 2.9 & 2.31 & 3.71   \\ 
FDR-PC         & -0.34 &	-0.43 &	0.55 &	-0.31	& -0.41	 & 0.51	& 0.34	& 0.41 &	0.53 \\
BDR-PC         & -0.15	& -0.16	& 0.22	& -0.04	& -0.04	& 0.06	& 0.14	& 0.16	& 0.21   \\ 
DDR-PC          &-0.11	&-0.14	&0.18	&-0.03	&-0.03	&0.04	&0.11	&0.14	&0.18   \\ 
\hline
\end{tabular}

\caption{
Displacement results at $\textbf{P}_2$ for various PCs under different load conditions.
}
\label{table:p2}
\end{table}

\subsection{Deformation Behavior of Door-Latch Mechanism}
\begin{figure}[h]
\centering
  \includegraphics[width=.8\textwidth]
  {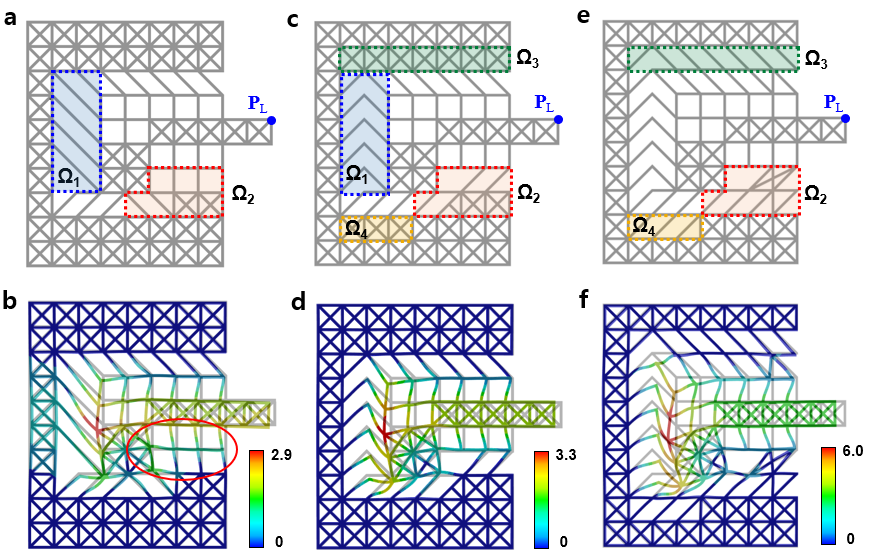}
\caption{FEA results for various door-latch mechanisms ($T=10$ N-m): (a) mechanism design (Design 1), (b) deformed shape (Design 1), (c) mechanism design (Design 2), (d) deformed shape (Design 2), (e) mechanism design (Design 3), and (f) deformed shape (Design 3).}  \label{fig:door_fea} 
\end{figure}

Figure~\ref{fig:door_fea} presents the deformed configurations of the door-latch structures for three distinct design cases outlined in Figure~\ref{fig:structure}c, with the color contour representing the displacement magnitude. Table~\ref{table:platch} compares the area densities of the designed structures and the resulting directional displacements at the latch tip ($\textbf{P}_\text{L}$) for rotational torques of 5 and 10 N-m, respectively. Overall, the data indicate that Design 3 yields the most favorable results, exhibiting the largest horizontal displacement ($u_x$) while maintaining minimal vertical displacement ($u_y$).

Figure~\ref{fig:door_fea}a depicts the first design case (Design 1), where the marked region ($\Omega_1$) incorporates several PCs on the left side of the axle. This design is based on the findings of the preceding section, highlighting the superior flexibility of PCs in shear deformation compared to SCs. Figure ~\ref{fig:door_fea}b displays the resulting deformed shape under a rotational torque of 10 N-m, with the color contour indicating the displacement magnitude. The deformed configuration reveals maximal deformation localized at the top-left corner of the axle, measuring 2.95 mm in magnitude. Notably, the inclined ligaments within $\Omega_1$ undergo stretching deformation, effectively moderating the rotational motion of the axle. The horizontal ($u_x$) and vertical ($u_y$) displacements at $\textbf{P}_\text{L}$ are measured at 1.64 and -0.67 mm, respectively, with the corresponding displacement slope ($u_y/u_x$) calculated to be -0.428. To meet the functional requirements of a door-latch mechanism, additional modifications are required to increase horizontal displacement while decreasing vertical displacement, thereby maintaining the displacement slope as small as possible.

To enhance the horizontal displacement, the configuration of ligaments within $\Omega_1$ has been modified to feature a $90^\circ$ bend, as shown in Figure~\ref{fig:door_fea}c. Additionally, the configuration of ligaments within $\Omega_2$ has been revised to incorporate pure PCs to reduce vertical displacement. This modified design, referred to as Design 2, is illustrated in its deformed state in Figure~\ref{fig:door_fea}d, demonstrating that the bending deformation occurs dominantly in the ligaments within $\Omega_1$. Consequently, the horizontal displacement of the latch increases to 2.06 mm, marking a 25.6\% improvement compared to the preceding design (Design 1). Moreover, the vertical displacement is reduced to 0.112 mm, and the corresponding displacement slope is significantly decreased to 0.055.

Figure~\ref{fig:door_fea}e illustrates a further modified design (Design 3) for additional improvement of compliance. Two near-boundary regions ($\Omega_3$ and $\Omega_4$) have been modified to include pure PCs instead of DDR-SCs, which exhibited the highest rigidity in the previous FEAs. In addition, the DDR-SCs in $\Omega_2$ are replaced with a pure PC and BDR-PC, while a DDR-PC is included to restrain vertical displacement. The resulting deformed shape is shown in Figure~\ref{fig:door_fea}f, indicating a significant increase in maximal axle deformation to 6.03 mm. As detailed in Table~\ref{table:platch}, the horizontal displacement at the latch tip increases to 3.24 mm, representing a 57.3\% improvement compared to that of Design 2 (2.06 mm). Moreover, the vertical displacement is notably reduced to 0.082 mm, with a corresponding displacement slope of 0.025. This outcome ensures the desirable movement of a door-latch mechanism characterized by substantial horizontal displacement with minimal vertical deflection.

\begin{table}[h]
\centering
\begin{tabular}{c|ccccc}
\hline
\multirow{2}{*}{Design case} & \multirow{2}{*}{Area density(\%)} &  \multicolumn{2}{c}{$\text{T} = 5 \text{ N-m}$} &  \multicolumn{2}{c}{$\text{T} = 10 \text{ N-m}$} \\ \cline{3-6}

 &  & $u_x$ (mm) & $u_y$ (mm) & $u_x$ (mm) & $u_y$ (mm) \\ 
\hline
Design 1 & 42.57 & 0.660 & -0.317 & 1.64 & -0.673 \\ 
Design 2 & 42.26 & 0.716 & 0.014 & 2.06 & 0.112 \\ 
Design 3 & 39.34 & 0.909 & 0.015 & 3.24 & 0.082 \\ 
\hline
\end{tabular}
\caption{Comparison of area densities and latch tip displacements for three designs of door-latch mechanisms (human-insighted designs).}
\label{table:platch}
\end{table}

\subsection{Deep RL for Optimal Design of Door-latch Mechanism}



\begin{figure}[h]
\centering
  \includegraphics[width=.8\textwidth]
  {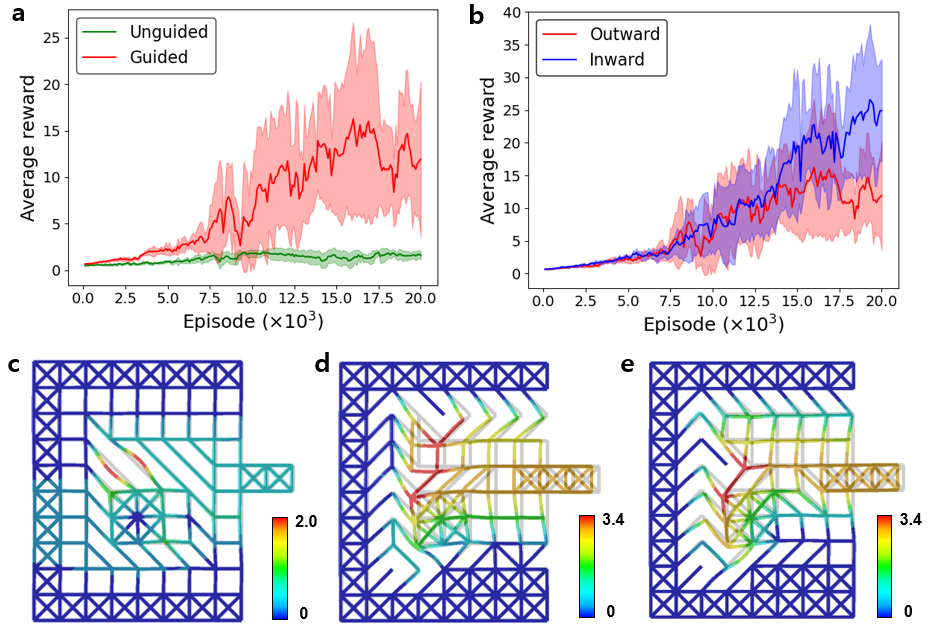}
\caption{Learning curves for the door-latch mechanism (a) effect of human guidance (outward tiling), (b) effect of tiling direction (guided), and the resulting door-latch structures with deformed shapes ($\text{T}=5$ N-m): (c) unguided outward tiling (Design 4), (d) guided outward tiling (Design 5), (e) guided inward tiling (Design 6).}  \label{fig:door:l}
\end{figure}

Figure~\ref{fig:door:l}a presents the learning curves of the door-latch mechanism based on human guidance in the initial configuration, when the tiling direction is set to the outward direction. It is evident that the initial configurations significantly affect the learning efficiency. Specifically, starting with the guided configuration consistently leads to a rapid increase in reward, with the human-guided configuration providing superior reward compared to the unguided configuration. Subsequently, the human-guided configuration is set as the initial configuration, and the RL procedure is conducted with different tiling directions. Figure~\ref{fig:door:l}b demonstrates the learning curves for both outward and inward tiling cases, revealing that learning through inward tiling outperforms outward tiling, as the former consistently achieves higher rewards.

The optimally designed door-latch mechanisms and their FEA results are shown in Figures~\ref{fig:door:l}c–~\ref{fig:door:l}e, corresponding to the unguided outward tiling (Design 4), guided outward tiling (Design 5), and guided inward tiling (Design 6), respectively. While the unguided design in Figure~\ref{fig:door:l}c exhibits local deformation near the axle with negligible latch displacement, the guided designs show significant horizontal displacements of the latch, as shown in Figures~\ref{fig:door:l}d and ~\ref{fig:door:l}e. Table~\ref{table:RL_latch} details the relevant displacements and reward values, including those from the human-insighted designs (Designs 1–3). Notably, Design 6 achieved the highest reward (28.39), which is three times higher than the best result from the human-insighted design (9.07 in Design 3). In this optimized design, the horizontal displacement is 2.84 mm, while the vertical deflection is as small as 0.003 mm. This outcome indicates that the proposed RL-based design, with appropriate human guidance and tiling direction, outperforms conventional mechanism designs depending on human insights.

\begin{table}[h]
\centering
\begin{tabular}{c|c|ccc}
\hline

Design no. & Description &  $u_x$ (mm)  & $u_y$ (mm) & \makecell[r]{$\text{Reward}$}\\ \hline
Design 1   &\makecell[l]{Human-insighted design}&  $0.660$     & \makecell[r]{$-0.317$}  & \makecell[r]{$3.29$}   \\ 
Design 2   &\makecell[l]{Human-insighted design}&  $0.716$     & \makecell[r]{$0.014$}   & \makecell[r]{$7.15$}   \\                 
Design 3   &\makecell[l]{Human-insighted design}&  $0.909$     & \makecell[r]{$0.015$}   & \makecell[r]{$9.07$}   \\ \hline
Design 4   &\makecell[l]{RL-based (unguided, outward tilling)}&  $0.468$     & \makecell[r]{$-0.096$}   & \makecell[r]{$4.29$}     \\ 
Design 5   &\makecell[l]{RL-based (guided, outward tilling)}&  $2.725$     & \makecell[r]{$0.059$}  & \makecell[r]{$26.33$}    \\
Design 6   &\makecell[l]{RL-based (guided, inward tilling)}&  $2.839$     & \makecell[r]{$0.003$}   & \makecell[r]{$28.39$}    \\
\hline
\end{tabular}
\caption{
Comparison of deformation results for various door-latch designs
}
\label{table:RL_latch}
\end{table}

\begin{figure}[h]
  \centering
  \includegraphics[width=1\textwidth]{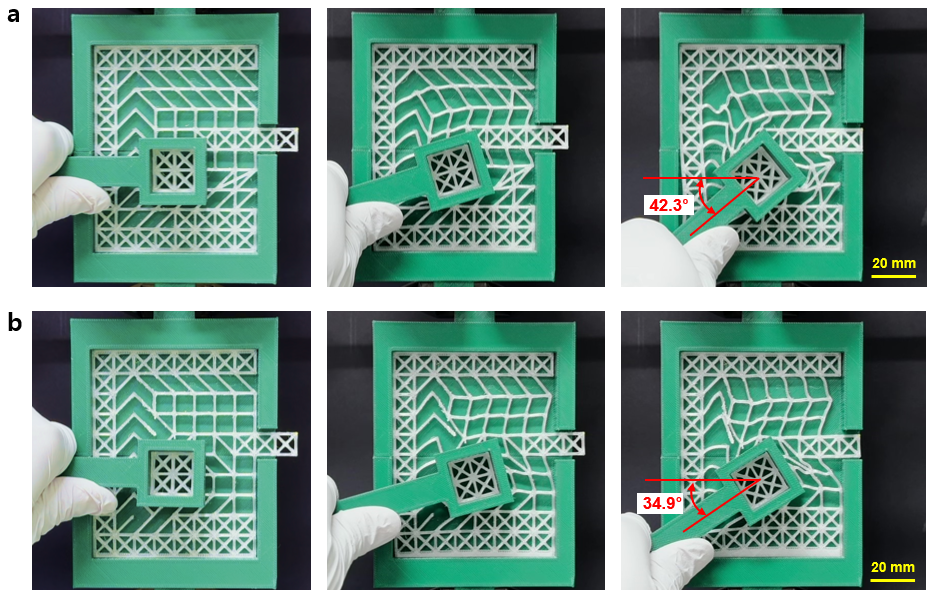}
  \caption{Stepwise deformed shapes of additively manufactured door-latch mechanisms: (a) human-insighted design (Design 3) and optimized design with RL (Design 6).}
  \label{fig:door_exp}
\end{figure}

\subsection{Experimental validation}
For experimental validation, the optimally designed door-latch structure (Design 6) is additively manufactured and assembled with the frame and handle, as described in Section~\ref{subsec:exp}. To compare the deformation behavior of the optimized design with that of a human-insighted design, AM is also performed for Design 3, which demonstrated the best performance among three human-insighted designs.

Figures~\ref{fig:door_exp}a and~\ref{fig:door_exp}b display stepwise deformations of Design 3 and Design 6, respectively, as the handle is rotated until the latch moves into the frame. This state corresponds to the door-opening condition, where the horizontal displacement of the latch reaches 10 mm. At this stage, the corresponding rotation angle of Design 3 is measured to be 42.3$^\circ$, while that of Design 6 is reduced to 34.9$^\circ$. This reduced rotation angle indicates that the compliant mechanism in Design 6 is more flexible than that of Design 3, thereby providing enhanced functionality as a door-latch mechanism. These findings highlight the effectiveness of the optimized design using deep RL in achieving more efficient compliant mechanisms. 

\section{Discussion}
\subsection{Design of Compliant Gripper Mechanism}
\begin{figure}[h]
\centering
  \includegraphics[width=0.9\textwidth]
  {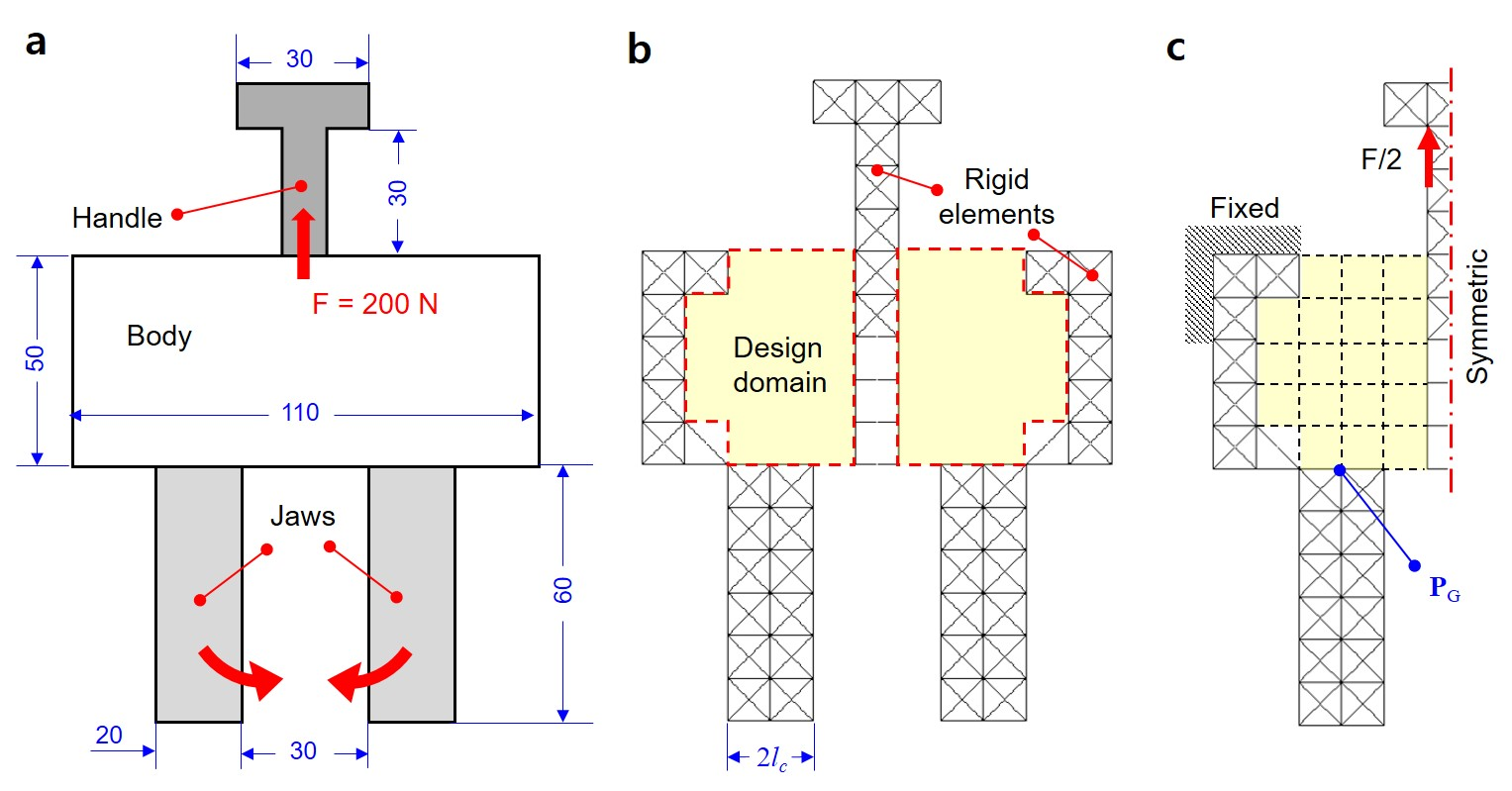}
\caption{Design of a compliant gripper mechanism: (a) basic design configuration (unit: mm), (b) initial configuration of design domain, and (c) FEA domain with boundary conditions.}  \label{fig:gripper_schema}    
\end{figure}

A compliant gripper mechanism is considered as an extended example to validate the proposed design methodology using deep RL. Figure~\ref{fig:gripper_schema}a shows the configuration of the compliant gripper mechanism, consisting of a handle, a body, and a pair of jaws. This gripper mechanism aims to induce lateral displacement of the jaws in response to the vertical displacement of the handle; upward movement of the handle causes inward motion for grasping, while downward movement induces outward motion for release. Each jaw has a size of $20 \times 60$ mm and is placed 30 mm apart. The body region measures $110 \times 50$ mm, where a finite number of cells will be designed to realize the compliant mechanism.

Figure~\ref{fig:gripper_schema}b presents the design domain of the compliant gripper mechanism. The non-design domains, specifically the handle and jaw regions, are arranged with a series of DDR-SC cells, as these regions are treated as rigid in the mechanism. The body region is then discretized into unit cells, each with a size of 10 mm. To ensure symmetric movements of the gripper, human-guided designs are applied to the center and two edge regions, as depicted in Figure 12b. This guided design approach helps maintain balance and uniformity in the gripper's operation.

The remaining region corresponds to the design domain, which consists of 18 cells, as indicated by the dashed line in Figure~\ref{fig:gripper_schema}c. Various cell types will be assigned within this domain to induce the desired compliant motion. An upward force of 100 N is applied to the handle considering symmetry, and the response of this mechanism is evaluated by measuring the rotation angle ($\theta$) of the reference point ($\textbf{P}_\text{G}$).

\subsection{Finite Element Analysis}
FEAs are conducted to investigate the deformation behaviors of soft gripper mechanisms. Figure 11c illustrates the FE model and the boundary conditions. The FEAs are performed using ANSYS $\text{Workbench}^\text{®}$ software (ANSYS Inc., USA). The FE model is prepared for a half section considering symmetry and is discretized using a 1D beam element with a mesh size of 0.1 mm. An upward force of 100 N is applied to the handle, and the upper corner region is restrained, as displayed in Figure~\ref{fig:gripper_schema}c. This setup allows for the detailed analysis of the gripper's deformation response under the given loading conditions, providing insights into the performance of the compliant mechanism.

\begin{figure}[h]
\centering
  \includegraphics[width=.9\textwidth]
  {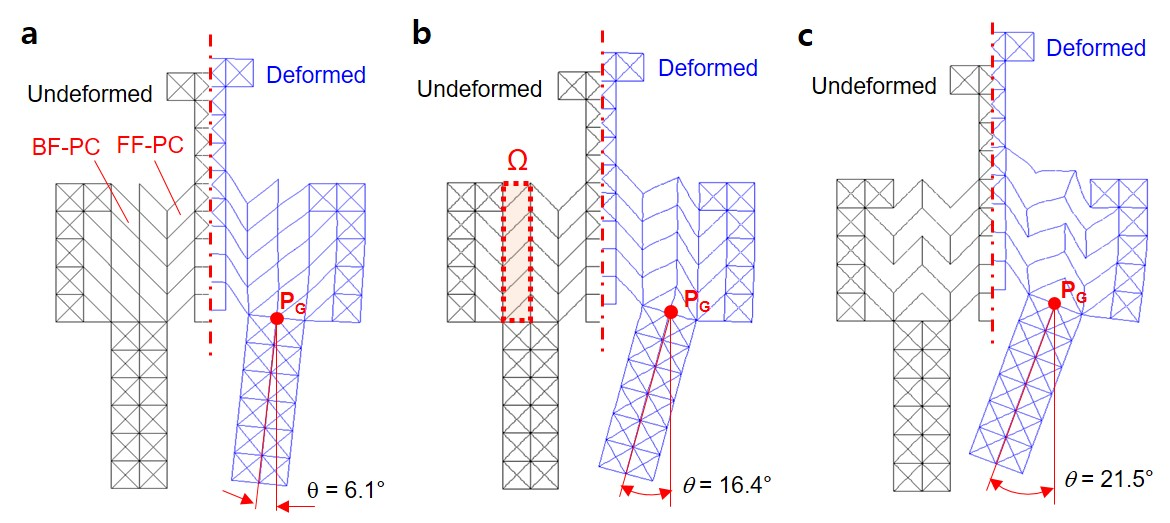}
\caption{FEA models and results for three gripper designs: (a) Design 1, (b) Design 2, and Design 3.}  \label{fig:gripper_FEA}    
\end{figure}

Figures~\ref{fig:gripper_FEA}a to \ref{fig:gripper_FEA}c present three design cases for soft gripper structures and their corresponding deformed shapes. Here, the undeformed and deformed geometries are displayed symmetrically with respect to the centerline in each case. Figure~\ref{fig:gripper_FEA}a depicts the first design candidate (Design 1), in which the design domain is filled with a set of PCs oriented in different directions. Specifically, the first vertical layer near the centerline is filled with forward-faced PCs (FF-PCs), while the other region is filled with backward-faced PCs (BF-PCs). This design, utilizing a set of pure PCs, is proposed based on the findings from Section~\ref{subsec:para}, which indicates that pure PCs exhibit high shearing compliance. The resulting rotation angle of the gripper (\(\theta\)) is then calculated to be 6.13$^\circ$.

Figure~\ref{fig:gripper_FEA}b illustrates the modified design (Design 2), where the BF-PCs in the marked region (\(\Omega\)) are replaced with FF-PCs. This design modification aims to alter the deformation mode of this region from shearing to bending by reorienting the direction of the PCs. As a result of this modification, the rotation angle at point $\textbf{P}_\text{G}$ increases to 16.39$^\circ$, corresponding to 2.67 times the rotation angle of the previous design (6.13$^\circ$). This outcome indicates that bending deformation is more advantageous than shear deformation for inducing the rotation of the gripper.

For further improvement of compliance, the design domain is modified by selectively removing vertical ligaments, as shown in Figure~\ref{fig:gripper_FEA}c. This design change aims to facilitate bending deformation by selectively removing vertical ligaments. As a result of this modification, the rotation angle ($\theta$) is increased further to 21.54$^\circ$. Notably, the gripper jaw deforms across the centerline, indicating that the two gripper legs are in contact with each other. Based on these empirical design insights, RL-based design optimization will subsequently be conducted to refine the compliant gripper mechanism, as described in the next subsection.

\subsection{Deep RL for Optimal Design of Gripper Mechanism}
The deep RL approach using the dueling DQN algorithm is employed to enhance the performance of the gripper mechanism. The reward function is designed to be proportional to the rotation angle (\(\theta\)). To stabilize the gripper mechanism, the number of disconnected hinges is included in the reward function as a penalization. This penalization aims to discourage disconnected hinges and promote learning a fully connected configuration, and the relevant reward function ($r$) is defined as:
\begin{equation}
r(s_t,a_t)=
\begin{cases}
    0\quad &\text{for} \quad t\in [1,H-1],\\ 
    \frac{C_1 \theta}{1 + C_2 d} \quad (\text{rad}) \quad &\text{for}\quad t=H,
\end{cases}
\end{equation}
where \(C_1\) and \(C_2\) are positive constants and \(d\) denotes the number of disconnections once the MDP reaches the terminal time. Similar to the door-latch case, $H$ denotes the horizon indicating the total number of cells to be placed. A hinge is defined as disconnected if it has fewer than two connections, which is used to penalize the design containing disconnected hinges. This penalization can be disabled by setting \(C_2 = 0\) when we want to disregard the effect of disconnection. Conversely, a reward with a large value of \(C_2\) will lead to a more stable mechanism design, as the reward function incentivizes the algorithm to learn a configuration with as many connections as possible. This reward function aims to find a configuration that induces the maximum rotation while maintaining the number of disconnections as small as possible.

\begin{figure}
\centering
  \includegraphics[width=0.9\textwidth]
  {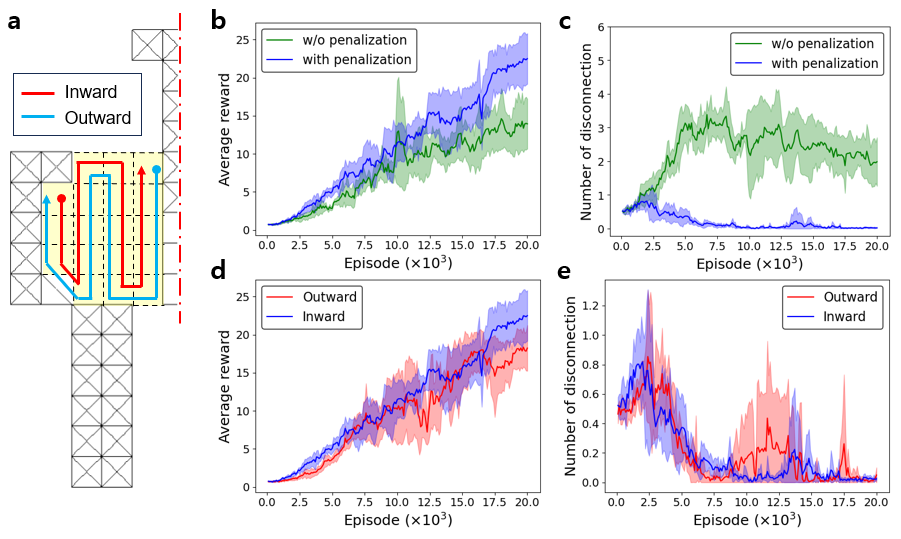}
\caption{Learning results for the gripper mechanism: (a) schematics of tiling direction, learning curves for different hinge-penalization methods (outward tiling): (b) average reward and (c) number of hinge disconnections, and learning curves for different tiling direction (with penalization):  (d) average reward and (e) number of disconnections.}  \label{fig:gripper_lc}    
\end{figure}

Figure~\ref{fig:gripper_lc}a demonstrates the initial configuration and tiling strategy for the half-gripper model. Inspired by the previous door-latch problem, human guidance is applied to the outer cells, and the remaining 18 cells are considered the reduced design domain. The vertical zigzag tiling is selected by accounting for the bending-dominated deformation of the gripper, with both outward and inward tiling directions considered. An upward force of 100 N is applied to the half model for FEA. All other hyperparameters are set the same as those used for the door-latch mechanism, with \(C_1 = 50\); \(C_2 = 0\) or \(1\) depending on the presence of hinge-penalization.

Firstly, the effect of hinge connection is investigated by selecting the outward tiling and varying the reward constant ($C_2$): $C_2=0$ implying no consideration of hinge-penalization, and $C_2=1$ implying the consideration of hinge-penalization. Figure~\ref{fig:gripper_lc}b presents the resulting learning curves for the reward value according to the penalization. Notably, the reward curve with penalization shows a higher increase than that without penalization. Figure~\ref{fig:gripper_lc}c illustrates the variations for the number of disconnected hinges, indicating that hinge penalization effectively reduces the number of disconnected hinges. Consequently, the penalized learning provides more efficient and stabilized learning as depicted in Figure~\ref{fig:gripper_lc}b.

\begin{figure}[h]
  \centering
  \includegraphics[width=.9\textwidth]{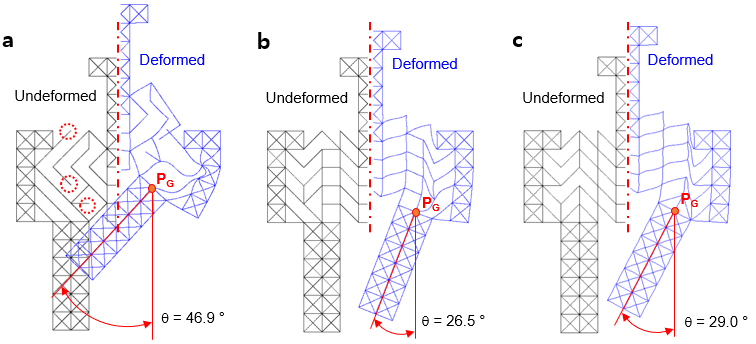}
  \caption{Optimally designed configurations with deformed shapes: (a) outward tiling without penalization (Design 4), (b) outward tiling with penalization (Design 5), (c) inward tiling with penalization (Design 6).}
  \label{fig:gripper_model}
\end{figure}

The effectiveness of hinge-penalization is verified through the FEAs for the RL-based designs, as presented in Figure~\ref{fig:gripper_model}. Figure~\ref{fig:gripper_model}a demonstrates the gripper model obtained from the unpenalized learning (Design 4) and its deformed shape, in which three disconnected hinges are observed. The resulting deformation exhibits an unstable shape, even though its rotation angle is as high as 46.9$^\circ$. In contrast, the penalized learning shows no disconnection and provides more stable deformation as demonstrated in Figure~\ref{fig:gripper_model}b (Design 5). While the resulting rotation angle, 26.5$^\circ$, is smaller than that of Design 4 (46.9$^\circ$), this value is larger than that of the best human-insighted design (21.5$^\circ$ in Design 3).

Another factor considered is the selection of the tiling direction, either outward or inward. Here, the penalizing constant ($C_2$) is set to 1 to impose hinge penalization. As a result, inward tiling slightly outperforms outward tiling, as demonstrated in Figure~\ref{fig:gripper_lc}d. The inward tiling also shows more stable results in the number of disconnected hinges, as depicted in Figure~\ref{fig:gripper_lc}e. The resulting design configuration for inward tiling (Design 6) and its deformed shape are provided in Figure~\ref{fig:gripper_model}c, showing an enhanced rotation angle (29.0$^\circ$) compared to that of the previous design with outward tiling (26.5$^\circ$ in Design 5).  

Table~\ref{table:gripper} compares the relevant rotation angles and reward values, including those from the human-insighted designs (Designs 1–3). Design 6 achieved the highest reward (25.3), which is 34.6\% higher than the best result from the human-insighted design (18.8 in Design 3). The resulting rotation angle is 29.0$^\circ$, showcasing a 34.9\% increase compared to that of Design 3 (21.5$^\circ$). This achievement demonstrates that the proposed RL-based design optimization outperforms human-dependent designs in terms of functional performance and structural stability, with an auxiliary penalization of hinge disconnection.

\begin{table}[h]
\centering
\begin{tabular}{c|c|cc}
\hline

Design no.   & Description&   $\theta$ (degree)  &  $\text{Reward}$\\ \hline
Design 1     &\makecell[l]{{Human-insighted design}}       & \makecell[r]{$6.1$}     & \makecell[r]{$5.35$}   \\ 
Design 2     &\makecell[l]{{Human-insighted design}}       & \makecell[r]{$16.4$}        & \makecell[r]{$14.30$}   \\                 
Design 3     &\makecell[l]{{Human-insighted design}}       & \makecell[r]{$21.5$}        & \makecell[r]{$18.80$}   \\  \hline
Design 4     &\makecell[l]{RL-based (w/o penalization, outward tilling)}   & \makecell[r]{$46.9$}     & \makecell[r]{$10.24$}     \\ 
Design 5     &\makecell[l]{RL-based (with penalization, outward tilling)}   & \makecell[r]{$26.5$}     & \makecell[r]{$23.10$}    \\
Design 6     &\makecell[l]{RL-based (with penalization, inward tilling)}   & \makecell[r]{$29.0$}     & \makecell[r]{$25.30$}    \\
\hline
\end{tabular}
\caption{
Comparison of deformation results for various gripper designs
}
\label{table:gripper}
\end{table}

\subsection{Experimental Validation}
\begin{figure}[h]
  \centering
  \includegraphics[width=.95\textwidth]{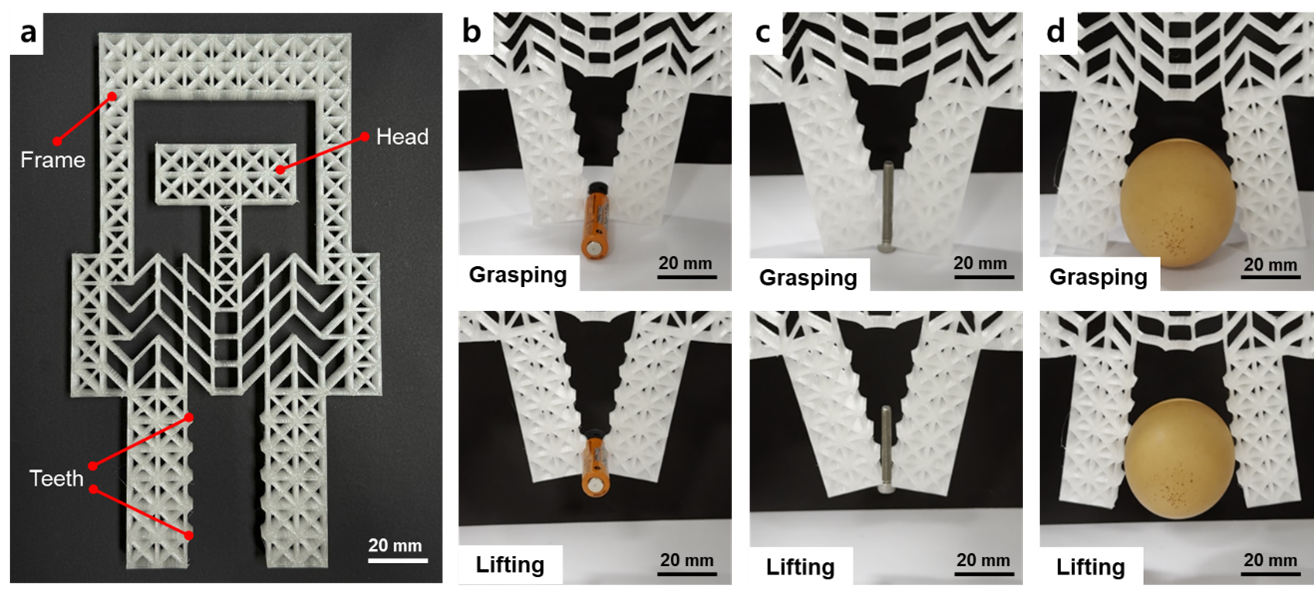}
  \caption{Experimental validations for compliant gripper mechanism: (a) a compliant gripper with additional features, and stepwise gripper motions for (b) an AAA-size battery (diameter: 10.3 mm), (c) a miniature screw (diameter: 4.0 mm), and (d) an egg (maximum diameter: 44.1 mm).}
  \label{fig:gripper_exp}
\end{figure}

For experimental validation of the optimally designed gripper structures (Figure~\ref{fig:gripper_model}c), additional features including a handle head and an auxiliary frame were designed as demonstrated in Figure~\ref{fig:gripper_exp}a. These additional features are intended to allow a human hand to push or pull the gripper handle, enabling the application of an upward or downward force on the integrated compliant gripper structure. Figure~\ref{fig:gripper_exp}a displays the additively manufactured gripper using TPU filament, which weighs only 77.2 grams due to its lightweight cell structure.

The additively manufactured compliant gripper is then validated by performing various gripper motions. Figure~\ref{fig:gripper_exp}b illustrates the stepwise motion of the gripper for an AAA-size battery as the target object. The AAA battery has a diameter of 10.3 mm, corresponding to one-third of the jaw distance (30 mm). As the gripper handle is pulled up, the two jaws rotate inward and grasp the target object. The lifting stage follows, where the handle position is maintained, demonstrating that the gripper can stably hold the battery. Finally, the gripper releases the target object by releasing the gripper handle, as can be found in the supplementary movie file (Movie 1). 

This sequential procedure, including the grasping and lifting stages, is repeated for a tiny object, a miniature screw with a diameter is 4.0 mm, as depicted in Figure~\ref{fig:gripper_exp}c. The entire motion, including the releasing stage, can be viewed in the supplementary movie file (Movie 2). These findings demonstrate that the proposed gripper mechanism operates successfully even for such a small object, as the proposed gripper design ensures high compliance with a maximized rotation angle of the jaws.

The compliant gripper mechanism is further tested by grasping a large object, an egg with a maximum diameter of 44.1 mm. Since this value exceeds the initial distance between the two jaws (30 mm), the gripper jaws must spread out to grasp the egg, unlike the previous cases. This reverse grasping motion can be achieved by pushing the gripper, as illustrated in Figure~\ref{fig:gripper_exp}d. The relevant grasping, lifting, and releasing stages are  successfully performed without fracturing the egg, as demonstrated in the supplementary movie file (Movie 3). This reverse operation underscores an advantage of the proposed compliant gripper, which can deform both inward and outward directions. Moreover, the compliant gripper mechanism is beneficial for handling fragile objects that might break under the operation of a rigid gripper.

\section{Conclusion}
In this study, we proposed an efficient and straightforward design methodology for compliant mechanisms based on deep RL. Design domains for compliant mechanisms were digitized into a finite number of cells with different hinge connections, Deep RLs were employed to obtain optimal compliant mechanisms that satisfy desired functional requirements by maximizing the reward in the sequential selection of cell designs. 

As a target mechanism, a compliant door-latch mechanism was designed to achieve a horizontal motion of a latch. Various designs from different combinations of digitized cells were analyzed using FEAs, the results of which were then effectively learned through a reward function. Deep RL using a dueling DQN framework was employed to determine the optimal compliant mechanism. The effects of human guidance and cell-tiling direction were investigated, revealing that the minimal human guidance and the inward tiling were more effective in achieving the desired motion. Consequently, the optimized design showed a three-times higher reward than a human-designed mechanism, resulting in 2.84 mm horizontal displacement with only 0.003 mm vertical displacement.

The proposed design approach was then extended to the design of a soft gripper mechanism, which initially resulted in unexpected hinge connections that deteriorated structural stability. The reward function was further refined to obtain the desired motion and avoid unconnected hinges. The effects of hinge connection and cell-tiling direction were investigated, revealing that hinge penalization and inward tiling were more effective in achieving the desired motion. Consequently, the optimized design showed a 34.6\% higher reward than a human design, resulting in 29.0$^\circ$ rotation angle. This optimized design was additively manufactured and validated through experimental gripping tests on objects of various sizes.

In summary, the proposed design approach, which involves obtaining an optimized combination of digitized cells through deep RL, offers an effective methodology for designing compliant mechanisms. Given that the optimized results demonstrated superior performances compared to human designs, the proposed approach not only enhances the efficiency of the design process but also ensures the creation of practical compliant mechanisms with less reliance on human expertise. Future studies are expected to extend the proposed approach to fully three-dimensional compliant mechanisms with more complex shapes.

\section*{Acknowledgement}
Keun Park is supported by 
the National Research Foundation of Korea (NRF) grant funded by the Ministry of Science and ICT (MSIT), Republic of Korea (2022R1A4A1032030), and Yeoneung Kim is supported by the National Research Foundation of Korea (NRF) grant funded by MSIT, Republic of Korea (RS-2023-00219980, RS-2023-00211503).

\section*{Data Availability}
All data that support the findings of this  study are included within the article (and any supplementary files).

\section*{Supplementary information}
\begin{itemize}
\item Movie 1: Grasping, lifting, and releasing motions for an AAA-size battery
\item Movie 2: Grasping, lifting, and releasing motions for a miniature screw
\item Movie 3: Grasping, lifting, and releasing motions for an egg
\end{itemize}

\bibliographystyle{elsarticle-num}
\bibliography{ref.bib}

\end{document}